# Temporal Data Requirement for Predicting Unplanned Hospital Readmissions

Ramin Mohammadi[1,2], Vahab vahdat[3,4,5], Sarthak Jain[1], Amir T. Namin[1], Ramya Palacholla[2,4,5], Sagar Kamarthi[1,2]

[1] Northeastern University, Boston, MA, US
[2] Partners Healthcare Connected Health Innovation, Boston, MA, USA
[3] MGH Institute for Technology Assessment, Boston, MA, USA
[4] Harvard Medical School, Boston, MA, USA
[5] Tufts University School of Medicine, Department of Public Health and Community Medicine, Boston, MA

## Abstract

With the advent of EHRs, one key challenge for building predictive models using EHR data is to determine an appropriate time-window of historical data to maximize predication accuracy. In this study, we investigate the impact of different observation windows for the prediction of 30-day readmission following hip and knee arthroplasties. We compare the performance of models built on different observation windows from the day-of-the-surgery to going as far back as 3 years from the surgery date. The patient data used in this study comprised of both unstructured (clinical notes) and structured formats. The data included over 4 million encounter records and 80,000 clinical notes of 7,174 patients undergoing one or more clinical procedures. We employed multiple non-neural encoders (Bag of Word-BOW, count BOW, TF-IDF, LDA) and neural encoders (BERT, 1D-CNN, BiLSTM, Average) to extract the meaningful information from the clinical notes. We developed a suite of models using the information extracted from clinical notes alone, structured data alone, and both. The results of this study indicate that the optimal time window for the clinical notes is generally shorter than that for structured data. When using clinical notes alone, the data within a time window of three to six months from the date of the surgery provided the best predictive performance. In contrast, when using structured data, the model performance improved with an increase in the length of the time window but plateaued when the time window exceeded beyond twelve months. We observed these patterns irrespective of the complexity of the model − for both non-neural encoders and neural encoders. These findings inform the length of the past historical data to be considered for optimal predictive performance for 30-day readmission prediction models as opposed to the general belief that the more the data, the better the machine learning model's predictive performance.

## Introduction

Access to a rich patient data set has become more achievable in the past two decades. Electronic Health Record (EHR) data, clinical notes, medical imaging, inpatient health monitoring, laboratory, and biometrics are a few examples of data-sources in healthcare. With the introduction of the Health Information Technology Act legislation in 2009[1,2], the use of EHR provided an impetus for developing data-driven predictive models.

Improving outcome of care, predicting risk of disease, and enabling precision medicine are among objectives of machine learning (ML) and statistical models that have been afforded by the abundant availability of patient data, which is estimated to be around 150 exabytes in 2016[3].

Previous studies employed historical EHR data, spanning over several years, to train data-driven models and these models typically used data from a long observation period to predict outcomes[4-6]. The observation window is the historic period during which the measurements or entries of independent variables /predictors are used for data analysis. Typically, researchers tend to use long observation windows to ensure adequate accuracy and reliability of the models they develop[5], but this work investigates if there are conditions that challenge this basis for structured and unstructured datasets. Structured data refers to datasets that are pre-identified, tabular, and formatted properly[4]. Models with structured data generally require the features to be manually engineered by domain experts and are limited by the size of the data[6]. However, clinical data also include a large volume of unformatted data, denoted as unstructured data, such as clinical notes, reports, and bio-signal data[7]. Studies show that 80% of healthcare data is unstructured and remains untapped[8]. The existence of such high-volume unstructured data has propelled the use of more complex methods to gain valuable insights from this data. Due to complexities in storing and analyzing the high-volume unstructured data, it tends to be abandoned in most medical centers[9]. Just as with structured data, little is known about the amount of historical unstructured data required to build reliable models.

We investigated the role of the length of observation window of historical EHR data on the predictive performance of ML models. Our objective was to identify the appropriate time window for historical data that generates and provides the best predictive performance with datasets containing unstructured, structured, and a combination of both.

We compared the performance of machine learning models that predict the risk of 30-day unplanned readmission in patients who have undergone hip and knee replacement procedures. We explored observation window lengths ranging from one day (that only includes the data from the day of surgery) to three years. We observed that utilizing more historical data for individuals did not necessarily lead to a better prediction performance when using unstructured data alone. A shorter observation period for models using unstructured data not only improves the predictive performance for 30-day unplanned readmissions but also decreases the need for storing, retrieving, and processing large amount of data.

## Methods

## Dataset

We collected retrospective data of 10,534 patients from their Electronic Health Records (EHR) with approval from the Institutional Research Board (protocol number 2016P002062 at Partners Healthcare). Patients included were adults aged 18 years or older who were admitted for hip or knee surgery in inpatient or outpatient settings, between the year 2006 and 2016. The Current Procedural Terminology (CPT) codes were used to identify patients who underwent hip arthroplasty (CPT Codes: 27130, 27132, 27134, 27236, 27137, 27138, 27120, 27125) and/or knee arthroplasty (CPT Codes: 27445, 27446, 27447, 27486, 27487). From the cohort of 10,534 patients, we excluded patients who were either deceased at the time of the study or older than 90 years at the time of surgery. It yielded 9,404 patients. Finally, patients that didn't have any clinical note in the EHR were dropped from the cohort, which reduced data to 7,174 patients.

For the consolidated patient cohort, we collected and analyzed structured and clinical notes consisting of over 4M encounter records, 15K clinical procedures, and 80K clinical notes (including over15k procedure notes). The structured data contained information in eight domains with different number of features, as highlighted in Table 1. The most detailed feature set was admission information with 25 features, followed by demographic information and health history.

Table 1. Number of variables collated by data categories

| Feature Category | Number of Variables | Sample set of features |
|---|---|---|
| Demographic information | 10 | Age, sex, education, language, race, marital status, religious, veteran, country |
| Laboratory test results | 4 | Lab test code, number, results and units |
| Diagnosis | 3 | ICD codes (code, results and units) |
| Medication information | 4 | Medication name, code, dosage, units, refill |
| Health history | 5 | Medical examinations (name, code, code type, results, units) |
| Procedure | 4 | Type and number of diagnosis (ICD 9, ICD 10 and DRG), type and number of procedures (CPT codes) |
| Vital information | 3 | Medical examinations (code, results, units) |
| Admission information | 25 | Admission date, discharge date, inpatient/outpatient, site location, length of stay, encounter status, service line, admit source, discharge disposition, payer, DRG, principal diagnosis, admitting diagnosis, type and number of diagnosis (ICD 9, ICD 10 and DRG), |

The primary source of unstructured data were clinical notes written by clinicians including physicians, nurses, and healthcare providers. In addition to the free text, the notes included semi-automatic and script-generated tables such as list of lab tests. All the notes were parsed to extract any information pertaining to the patient's demographic, procedures, medication, and other medical services.

We defined unplanned readmissions as patients who underwent hip or knee surgery and were subsequently readmitted to inpatient wards within 30-days after the surgery (represented positive class or 1). We assumed all other patients were not readmitted due to surgery complications (represented negative class or 0).

Out of 7174 (Hip: Male=1641, Female=1658; Knee: Male=1702, Female=2173) patients, the prevalence of positive class for hip replacement procedure was 0.095% and for knee replacement procedure was 0.105%. Since only a small portion of patients have been readmitted as inpatient, inherent class imbalance exists in the dataset[27]. We tested multiple strategies to counter class imbalance using class weights, undersampling, and oversampling.

For developing machine learning models, we partitioned data at the patient level considering training (70%), testing (15%) and validation (15%). In each of the partitioned dataset, patients' demographic remained consistent, as reported in the Appendix (Table A.1). To explore the effect of the clinical notes and the length of historical data on the performance of machine learning models, the following combination of the three datasets were used: a) clinical notes, b) structured data, and c) a combination of structured and clinical notes, as shown in Figure 1.

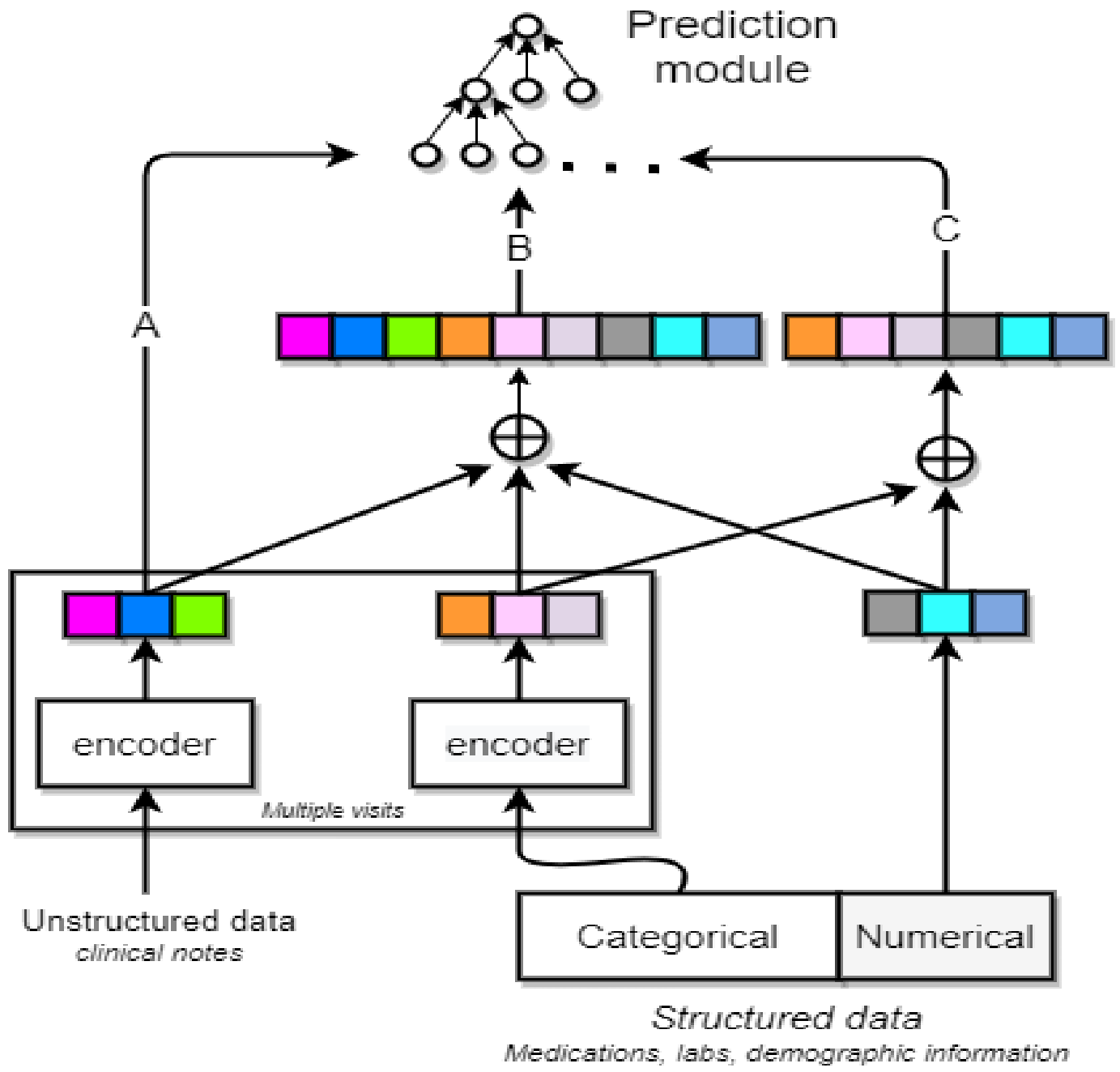

Figure 1. A schematic of feature encoding. Structured data contained both categorical and numerical elements. We encoded the categorical elements either via indicators or using an encoder module, while we packed the latter into a dense vector of values. Clinical notes (i.e., textual notes) are encoded using a sparse (indicator) representation and then optionally run through an encoder module.

## Model development

Figure 2 demonstrates a scheme of the developed models with different data sources. It shows the steps and flow of the data to neural and non-neural encoders. The details of the methods used for extraction and encoding of the data, processing structured and unstructured data, and model developments are described in this section.

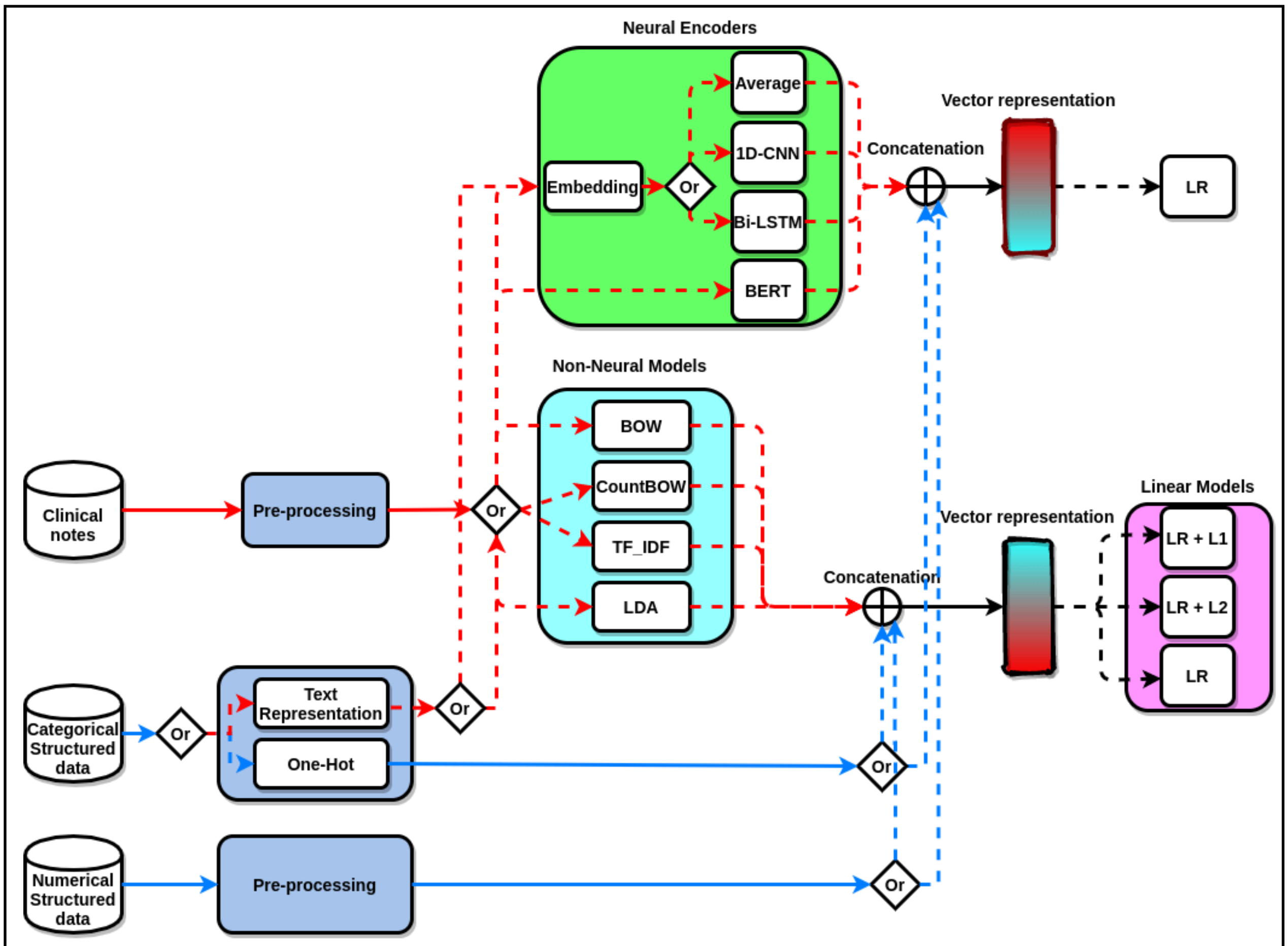

Figure 2. A schematic of model development, representing three sources of data, clinical notes, categorical, and numerical structured data. Flow of clinical notes are shown in red lines, while structured data flow represented in blue lines. Solid lines indicate signals having one pathway and dashed lines indicate the possible alternate pathways leading to model training. Diamond-shaped boxes are used to show alternative branches of data flow. 1D-CNN: One Dimensional-Convolutional Neural Networks, Bi-LSTM: Bi-directional Long Short Term Memory, BERT: Bidirectional Encoder Representations from Transformers, BOW: Bag of Words, TF-IDF: Term Frequency–Inverse Document Frequency, LDA: Latent Dirichlet Allocation, LR: Logistic Regression, LR+L1: Logistic Regression with L1 Regularization, LR+L2: Logistic Regression with L2 Regularization.

*Data Extraction and Encoding*

Since our predictive models are designed to work directly from extracted EHR, minimal pre-processing is required to convert the raw EHR data into vectors. The mapping process is automated unlike the manually defined predictors used in prior studies[28-30]. Details of pre-processing steps for structured and unstructured data are explained in Appendix.

## Independent Model Development

As shown in Figure 2, clinical notes were encoded into a vector representation to make it suitable for ML model building. We experimented with a few encoder variants: standard linear encoding methods (also called non-neural methods) such as unstructured count-based bag-of-words (BoW[32]) representations; and neural network encoders that operate over word embeddings and learn to represent notes via projection, recurrent, or convolutional modules. We explored the combinations of structured and clinical notes for both model types. For each of encoder types, we explored different text encoder options for inducing representations of the available clinical notes.

*Linear Models (Over BoW):*

We experimented with four different BoW variants: Binary BoW, which encodes the existence of a given word in a clinical note; Count BoW, which encodes the total number of occurrences of a given word in a clinical note; Term Frequency/Inverse Document

Frequency (TF-IDF)[33], which scales word counts inversely to the frequency of their appearance within the documents; and, Latent Dirichlet Allocation (LDA)[34] based text encoding via inferred 'topic' distributions.. In the last variant we encoded texts as vectors that determine the proportions of (latent) topics present within them, as estimated via LDA. As shown in Figure 2, we used standard Logistic Regression (LR); $L_1$-regulized LR, and $L_2$-regularized LR over each non-neural representations of the patient notes and the structured data to predict readmission.

## *Neural Encoders*

We also explored various neural NLP architectures for text encoding, as shown in Figure 4. This entails first projecting words to lower dimensional embeddings, which is in our case 300 dimensions (note that these may be initialized to pretrained embeddings); it yielded a matrix $E \in R^{TL \times E}$. We passed the embedded input $E$ through an upstream ***encoder*** module prior to making a prediction. We considered the following options for the encoder component.

**Average:** It involves projecting and then averaging inputs[35]. Specifically, we first passed embeddings through a linear layer that projects them into a 256-dimensional space, and then applied an element-wise non-linearity (ReLU) function.

**Convolutional Neural Network (CNN)[36]:** In this approach we performed 1D convolutions on the embedded sequence using different kernel sizes (3, 5, 7 and 9) using 64 filters for each size (256 overall).

**Bidirectional Long Short-Term Memory (BiLSTM) network[37]:** In this approach we ran a single layer bi-directional LSTM model over the embedded sequence, using a hidden size of 256 (128 dimensions for each direction).

These encoders provided outputs with dimensions varying with the size of the input; therefore, the encoders outputs were aggregated into fixed-dimensional vectors by using different aggregation strategies.

We adopted a standard max-pooling layer for CNNs and RNNs over the outputs from the 256 filters/hidden units. We explored aggregation through the attention mechanisms[38], which enable models to look at individual hidden descriptions of inputs according to greater influence over the induced fixed-length vector that passed to downstream layers. Specifically, the attention mechanisms learned to aggregate encoder outputs. In the standard attention layer, the model learned to score each hidden state $h_t$ induced by an encoder for input token $t$ in accordance with its relevance for the downstream prediction. These scores were normalized into a distribution $\alpha$ and a fixed length vector for downstream consumption which then induced by taking a weighted sum $\sum_i \alpha_i h_i$.

In addition, we experimented with hierarchical representation learning over notes[39]. We passed a single BiLSTM to embed each note separately (using attention mechanism), and then we ran another BiLSTM over the aggregated representation of the individual notes (associated with its own attention distribution). This aggregated representation passed forward to downstream modules for prediction.

Finally, we presented preliminary results using Bidirectional Encoder Representations from Transformers (BERT)[40] as another encoding strategy. Specifically, we used the clinical BERT[41] instantiation of the model that was trained on clinical notes from MIMIC III dataset. BERT is a deep bidirectional model that conditions on both left and right context at each word in the text to provide contextualized representations. It is the state-of-the-art model on multiple standard NLP tasks and has shown improvement for 30-day readmission tasks using MIMIC dataset.

## Multitask Learning

To handle the two predictive tasks appropriately, 30-day readmission following 1- hip and 2-knee arthroplasties, we constructed an independent model for each surgery type, considering as an entirely distinct task. In addition to the independent models, we also evaluated multitask model variants that jointly predicted readmission due to complication following either hip or knee surgery. In *multitask learning*[42], a subset of parameters was shared between predictive modules for different but related tasks. This procedure shared the encoder components between the hip and knee surgery complication prediction models. Intuitively this approach allowed borrowing strength across tasks which may enhance the predictive performance.

### Generalization

Important need of the clinical data for ML-based solutions is no secret but how far should we go back in time to have a great accuracy is unknow. We hope that this initial effort inspires additional work on identifying the appropriate time window for historical data that generates and provides the best predictive performance with datasets containing unstructured, structured, and a combination of both.

## Results

The data of adults aged 18 years or older, who had undergone surgeries in inpatient or outpatient settings between 2006 and 216, were retrospectively collected for 10,534 hip and knee surgeries. To extract useful information from the clinical notes we implemented four different non-neural encoders including Binary Bag of Words (BOW), count BOW, Time Frequency-Inverse Document Frequency (TF-IDF), and Latent Dirichlet allocation (LDA) and various neural NLP architectures including Average,

One-dimensional Convolutional Neural Network (1D-CNN), Bi-directional Long Short Term Memory (Bi-LSTM), and Bidirectional Encoder Representations from Transformers (BERT). We then designed and developed individual and multitask models based on the surgery type (hip or knee) to predict the 30-day readmission based on the extracted features from the clinical notes, structured data, and a combination of the clinical notes and structured data.

To evaluate the performance of the model variants, Area Under the Receiver Operating Characteristic (AUROC) curve is used. Figure 3 shows the average AUROC and the 95% confidence interval from all the model variants for knee and hip readmissions. The results measured the performance of the independent prediction models for the validation set, using (i) clinical notes (ii) structured data and (iii) combination of clinical notes and structured data.

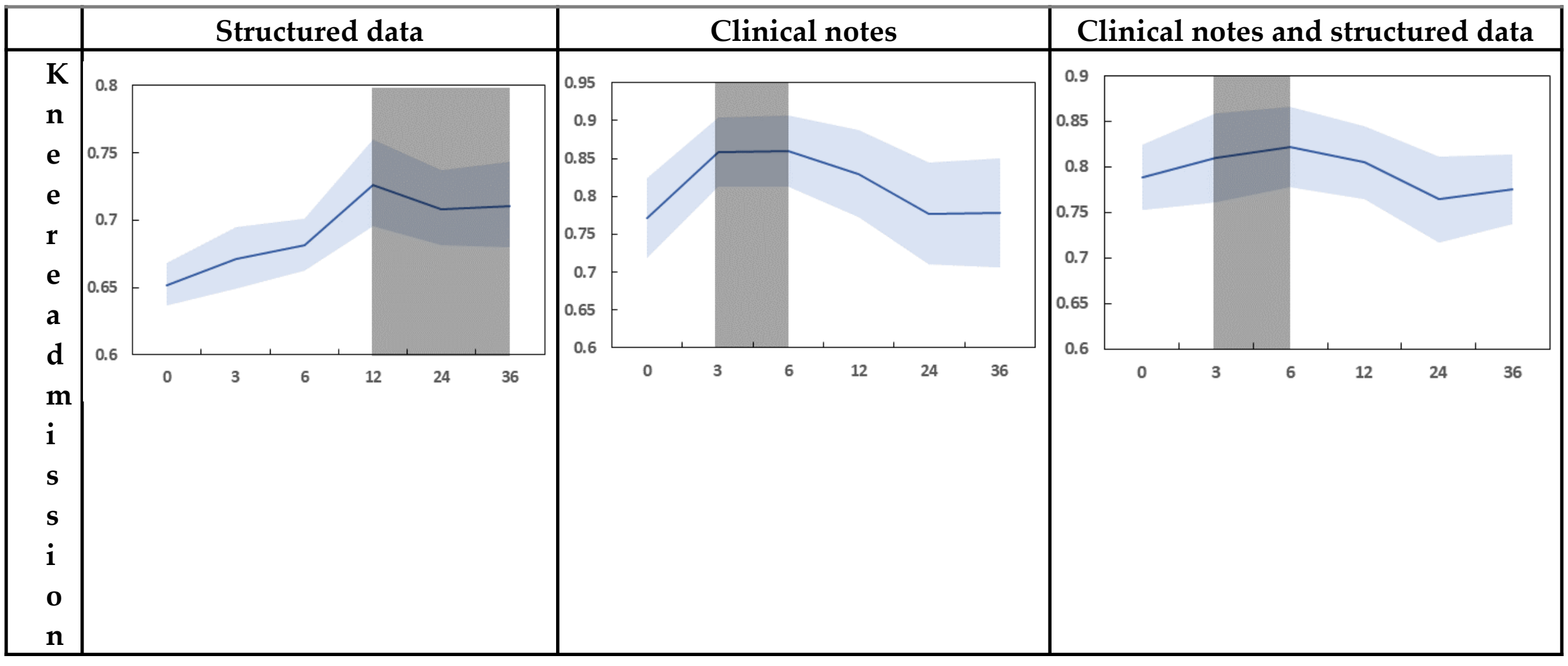

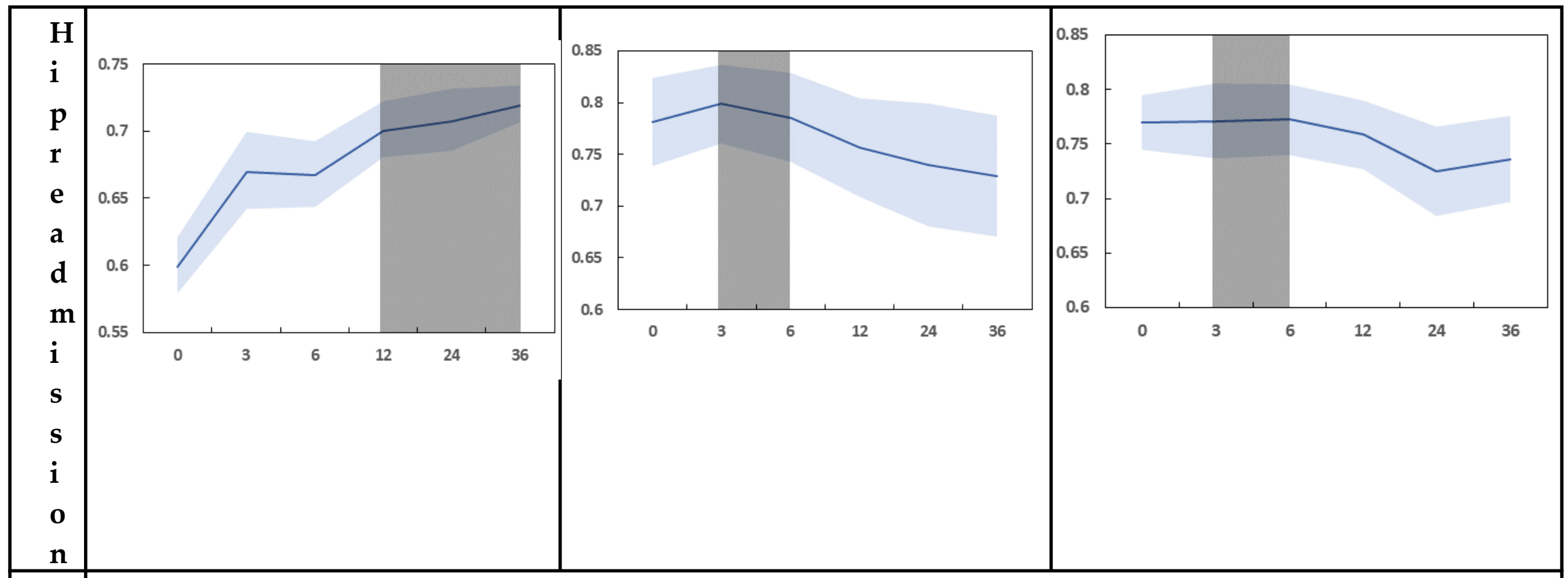

Figure 3. The variation in performance (mean and 95% confidence interval) of AUROC (*y-axis*) for prediction of knee and hip readmission with respect to the duration of the historical data. *x*-axis represent the length of observation window of training data; the observation window varies from the day of surgery (treated as 0-month prior information) to three years past the surgery (36-months prior information). Blue line represents the average AUROC of all model variants and the band shows the 95% confidence intervals.

Details regarding the performance of each individual model with the corresponding data encoding for both hip and knee surgeries are presented in the appendix (Tables A.2 to A.7).

As shown in Figure 3, the pattern of AUROC with temporal data requirements remain the same across hip and knee surgeries. Our results show that the pattern of temporal data requirements for clinical notes are different from those for structured data. When using only structured data, the performance of the models kept improving up to 12 months of historical data, but the performance plateaued after 12 months. When using only clinical notes, the best performance was achieved for 3 to 6 months period of the historical data prior to the surgery. When using both structured data and clinical notes together, the performance pattern was similar to case of using only clinical notes. This indicates the superiority of clinical notes over structured date when it comes predictive performance.

Other alternative models were constructed to perform multiple tasks independently or jointly. The models with independent tasks handle decisions for each surgery type individually and the multitask learning models exploit structure and similarities across both surgeries. While in many healthcare applications, multitask learning has led to performance improvement compared to models learning independent task, we were interested to verify the historical data requirements in prediction of readmission surgery for such models. For this purpose, we further verified the performance of independent and multi-task models over the validation and test datasets. Figure 4 presents the trend of historical data requirement for such models, categorized by type of data being utilized. For each observation window, all the model variants are explored and the highest performance model in response to each independent and multitask learner are shown in Figure 4. The results indicate that multitask learners may give better performance when both clinical notes and structured data are used. However, the trends of historical data requirement are consistent across independent and multitask learners, indicating that the data requirement is implicitly independent of model variants and multitask learning.

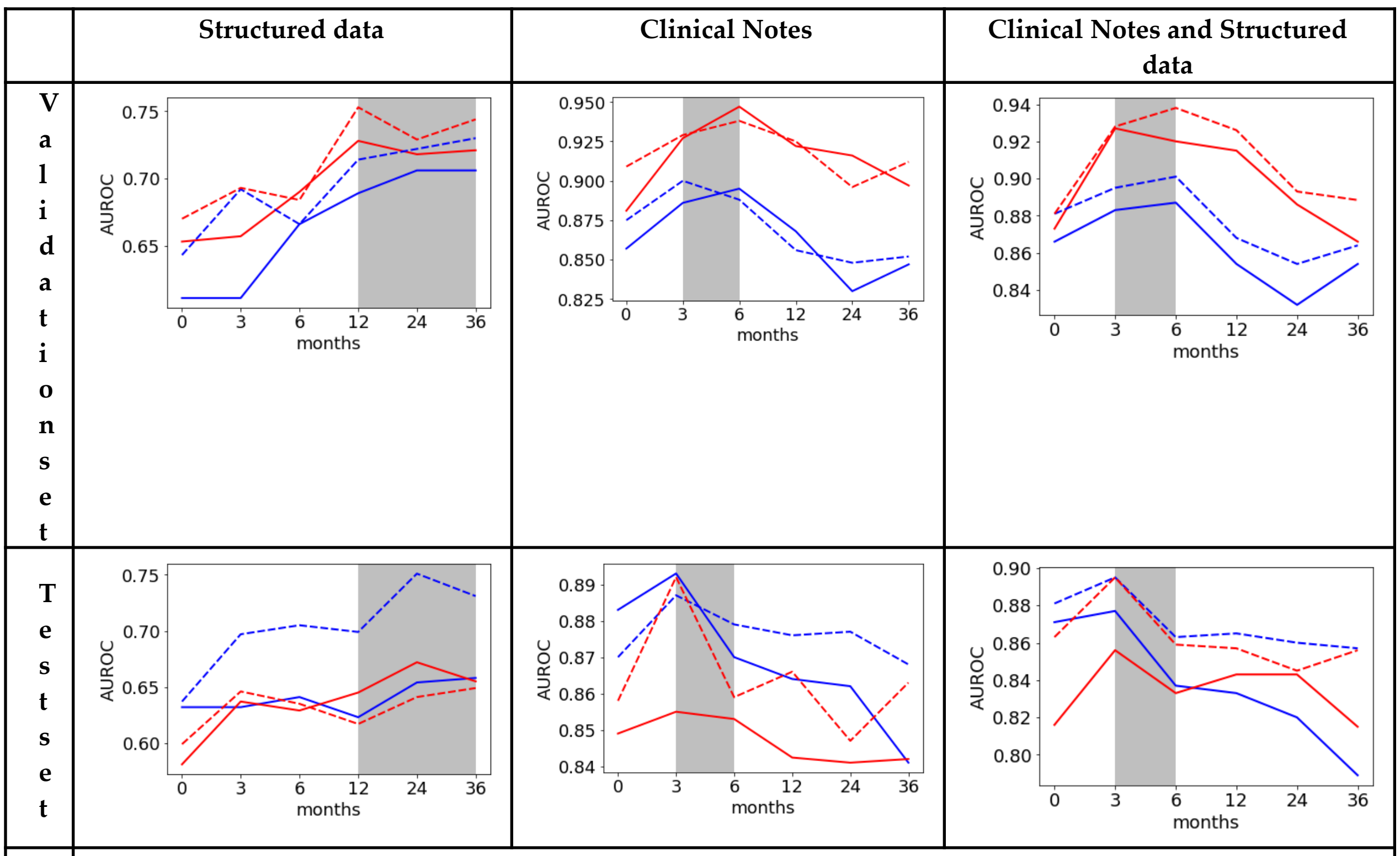

Figure 4. AUROC of the four models for prediction of knee and hip readmission versus observation window length. Blue lines represent hip prediction models and red lines denote the knee prediction models. Solid-colored lines show the independent-task models and dashed lines show multitask models. $x$-axis represents the length of observation window of training data and it varies from day of surgery (0-month prior information) to three years past the surgery (36-months prior information).

## Discussion

Unplanned readmissions consume hospital resources that can otherwise be deployed for patients in need. In the US, insurance companies penalize unplanned hospital readmissions for selected procedures[10] . In Massachusetts, for example,

Medicare penalized 78% of hospitals for unplanned readmissions between 2015 and 2016[11]. Predicting unplanned readmissions after a surgery can help utilize healthcare resources effectively and minimize regulatory punitive burden.

For the last two decades, Machine Learning (ML) models have been explored for predicting patient readmissions from risk of complications following surgery dates[12]. However, previously published models did not investigate how the length of the historical data affects the predictive performance of the models. A research found that models built on only the last visit information achieved lower accuracy than the models built on all the past readmission history, although authors did not investigate how much of the past is optimal for the readmission models[13]. In one study, data over one year from January 2013 to December 2013 were utilized for analysis of total joint arthroplasty readmissions[14]. Among models that used longer historical data, a study of two years data from 2017 to 2019 showed the association between infection and readmissions, albeit no explanation for observation period is provided[15]. Thus, there is a need to investigate the optimal duration required for the historical data that can provide the highest predictive performance for a given model.

Not only has the length of observation period been previously overlooked, but also there is no agreement in the types of data (i.e. surgery notes, structured data, clinical notes) to be utilized for 30-day readmission. Possibility of performance improvement for 30-day readmission prediction with unstructured data by using larger datasets is demonstrated previously[16]. However, another study speculated that longer length of observation period would not necessarily improve the predictive performance and it utilized free-text data over a period of two months for prediction[17].

We found that the pattern of longitudinal data requirements is consistent for different machine learning models across both types of surgeries. In previous studies, structured historical data for inpatient clinical order recommender system showed the similar decay patterns after 4 months[18]. They found that prioritizing a small amount of current data is more effective than using larger amount of older data. However, the observation was limited only to structured data. As summarized in two systematic reviews[19,20], a majority of the data-driven models that predicted surgical site infections and 30-day readmissions (e.g.[21,22]) demonstrating moderate success, were based on structured data. The effectiveness of combined use of structured and unstructured data is demonstrated previously[6]; and our study provides important evidence on leveraging a combination of historical clinical notes and structured data on the prediction of 30-day readmissions.

The study findings also highlight the importance of including patient's medical history, in addition to surgery notes, for predicting unplanned readmission. We studied the usefulness of two sources of the data for unplanned readmission prediction: (1), unstructured and structured data generated of the day of surgery including variables reported from previous studies [23-25], and (2) unstructured and structured data generated on the day of the surgery and in the past 3 months (we only took 3 months of past data considering the results reported in A-4 through A-9 that 3 months is the optimal for unplanned readmission prediction). Table 2 shows the performance of best classifiers for each surgery type and data sources type. Our findings show that the clinical texts from the day of surgery when it is combined with the 3 months of historical data give a better prediction than the day of the surgery data alone.

Table 2. The effect of surgical data on the model performance. The model without historical data that only utilizes surgical notes is compared with a model that only uses 36-month historical data (without surgical notes) and a model that uses both surgery notes and 3-month of data. Bold numbers show the model with highest performance.

| | **Hip Surgery** | | **Knee Surgery** | |
|---|---|---|---|---|
| | **The day of the Surgery data** | **The day of the Surgery data + 3 months historical data** | **The day of the Surgery data** | **The day of the Surgery data + 3 months historical data** |
| **Structured data** | 0.643 | **0.692** | 0.668 | **0.693** |
| **Clinical texts** | 0.875 | **0.890** | 0.909 | **0.929** |
| **Clinical texts + Structured data** | 0.881 | **0.888** | 0.881 | **0.930** |

## Limitations

There are inherent limitations associated with these types of studies. First, the opinions of experts such as surgeons and nurses regarding the patient's risk of complications are not considered in the evaluation process. These opinions can provide a strong baseline for predictive models. Second, personalized predictors such as patient's socio-demographics, personal genomics, mobile-sensor readouts, and other information streams that could have improved predictive performance[26], are not incorporated. The EHR data also imposes some limitations to this research. Imputation of missing values in addition to the sampling of patient EHR data may add biases to the predictions. The conversion of medication and lab results may cause some information loss, though we tried to reduce the information loss through domain knowledge gained in consultation with experts. In addition, manually formed sets of ICD codes to create label categories (complication/ no complication) may lead to unknown biases in the positive cases (patients with readmission). Finally, the data for this study comes from one specific EHR system (Epic); it might impose geographical or systemic bias in our finding.

For future analysis, models for readmission risk after other surgeries such as cardiovascular can be investigated. We are also interested in validating the findings on much broader data generated by different organizations.

## Data availability

Data supporting this study are not publicly available because of the inherently sensitive nature of the data.

## Appendix and Supporting Information

### A.1 Patient demographic information

| | | Training set | | Testing set | | Validation set | |
|---|---|---|---|---|---|---|---|
| | **Gender** | **Female** | **Male** | **Female** | **Male** | **Female** | **Male** |
| | **Total sample** | **2744 (55.32%)** | **2216 (44.68%)** | **555 (52.21%)** | **508 (47.79%)** | **563 (52.91%)** | **501 (47.09%)** |
| **Age** | **Min.** | 20 | 24 | 38 | 26 | 25 | 28 |
| | **1st Quartile** | 62 | 59 | 62 | 59 | 62 | 60 |
| | **Median** | 69 | 68 | 70 | 67.5 | 69 | 68 |
| | **Mean** | 68.92 | 66.82 | 69.55 | 66.54 | 69.12 | 67.09 |
| | **3rd Quartile** | 77 | 75 | 77 | 75 | 77 | 75 |
| | **Max.** | 89 | 89 | 89 | 89 | 89 | 89 |
| **Race** | **White** | 2353 (85.75%) | 2007 (90.05%) | 473 (85.22%) | 457 (89.96%) | 481 (85.43%) | 448 (89.42%) |
| | **Black or African American** | 107 (3.89%) | 54 (2.43%) | 25 (4.50%) | 9 (1.77%) | 31 (5.50%) | 16 (3.19%) |
| | **Asian** | 62 (2.25%) | 109 (4.91%) | 17 (3.06%) | 29 (5.70%) | 10 (1.77%) | 28 (5.58%) |
| | **Other** | 179 (6.52%) | 25 (1.12%) | 34 (6.12%) | 7 (1.37%) | 30 (5.32%) | 6 (1.19%) |
| | **Hispanic** | 43 (1.56%) | 21 (0.94%) | 6 (1.08%) | 6 (1.18%) | 11 (1.95%) | 6 (1.19%) |

Table A.1: Demographic of included patients in training, validation, and testing datasets. Age distribution and race proportions are retained similar among dataset partitions.

## A.2 Dropped Variables

**Variables dropped from consideration due to high proportion of missing values (> 99.9%):**

AAA (Abdominal Aortic Aneurysm ) Screening , ANA Screen , Albumin/creatinine ratio , Alcohol Drinks Per Week , Alcohol Oz Per Week , Alcohol Use Screening , Amino Acids , Amino Acids, urine , Antibody Screen , Antiphospholipid Antibodies , Antiphospholipid Antibody , Auto-Antibodies , B12 injection , Blood Gases/Oximetry , Blood Pressure-LFA1162 , Blood Type , Body Surface Area (BSA) , Bone Marrow Stain , Bone density , Breast Exam , Breast Exam - LHA3537 , Breast Exam - LHA4003 , Breast Exam Instruction , CRYOs , CSF Chemistries , CSF Counts and Diff , CSF/Fluid, Other , Calcium Requirements Recommendation , Carnitine, serum , Carnitine, urine , Chlamydia , Cholesterol , Cholesterol-HDL , Cholesterol-LDL , Cigarettes , Coagulation Factor Studies , Colonoscopy , Complement , Complete Physical Exam , Condoms , Creatinine , Cystic Fibrosis Carrier , DNA Diagnostic Tests , DPT , DS Glucose , Dental Exams , Depo-provera Shot , Diet , Diphtheria and Tetanus booster (DT booster) , Domestic Violence Screening , Drug Use Screening , Drugs A-E , Drugs F-N , Drugs O-Z , EGD (upper GI endoscopy) , EKG , Echocardiogram , Exercise Advice , FEV1-pre (Pre-Forced Expiratory Volume) , FVC-pre (Pre-Forced Vital Capacity) , Fetal Activity , Fluid Chemistries , Fluid Counts and Diff , Folic Acid Recommendation , Foot exam , Functional Status Screen , GFR (estimated) , Glucose , Gonorrhea , HCG (Human Chorionic Gonadotropin) , HCV Ab-LHA3507 , HIVx, Haemophilus Influenzae type B (HIB) , Hand Gun Counseling , HbA1c (Hemoglobin A1c) , Hct (Hematocrit) , Head Circumference , Hearing , Hemocult x 3 , Hemoglobin Electrophoresis , Hepatitis A vaccine (Hep A vac) , Hepatitis B vaccine (Hep B vac) , Hgb (Hemoglobin) , HgbAIC , Home Hemocult , Home glucose monitoring , Hypercoagulation Studies , Hypoglycemia Assessment/Counseling , INR Result , Immune globulin , Inhibitors , Japanese encephalitis , KPS (Karnofsky performance status) , Liver - AST , Liver - Alkaline Phosphatase , Liver - Total Bilirubin , Liver ALT , Lyme , Lyme vaccine , Lymph - % Difference , Lymph - Left Arm Volume , Lymph - Right Arm Volume , Mammogram , Measles, Mumps, Rubella (MMR) , Medicare Annual Wellness Visit , Meningococcal vaccine , Microalbumin , Nutrition Referral , O2 Saturation - LFA15000 , O2 Saturation - LFA15000.1 , O2 Saturation - LFA12575 , O2 Saturation - LFA38131 , O2 Saturation - LFA38132 , O2 Saturation - LFA4826 , O2 Saturation - LFA4828 , O2 Saturation - LFA5392 , O2 Saturation - SPO2 , OPV / IPV , On Oxygen? , Ophthalmology Exam , Organic Acids, urine , PSA , Pain 0-10 , Pain Assessment , Pain Scale (0-10) , Pain Score , Pap

Smear , Peak Flow , Peak Flow - LHA4483 , Pelvic Exam , Personal Best Peak Flow , Platelet Aggregation , Platelet Antibodies , Pneumovax , Podiatry exam , Positive Antibody Screen , Pregnancy Weight , Prepregnancy Height , Prepregnancy Weight , Principal ICD Procedure CD , Prostate exam , Rabies , Rabies immune globulin , Rapid Strep , Rectal Exam , Rh Factor , Routine Serology , Safe Sexual Practice Counseling , Seat belt counseling , Second hand smoke exposure , Sigmoidoscopy , Smoking Quit Date , Smoking Start Date , Special Coagulation Interp , Stool Guaiac - 3 , Stool Guaiac-LHA4072 , T-cell Subsets , TSH-LHA18009 , Testicular Exam , Testicular Exam Instruction , Tetanus, Diphtheria, accellular Pertussis vaccine , Tobacco Pack Per Day , Tobacco Used Years , Toxicology , Triglycerides , Trisomy 21 , Tuberculin purified protein derivative , Typhoid , UA-Protein , Urine Chemistries , Urine Chemistries Timed , Urine Chemistries Unspec , Urine Culture , Urine Dip-LHA4935 , Urine Glucose , Urine Protein , Urine Toxicology , VAS score , Varicella , Vision , Vision-Left Eye , Vision-Right Eye , Vitamin D (25 OH) , Weight Management , Yellow fever

## A.3 Surgical Texts Example:

### *A.3.1 Surgery with Complication:*

NAME OF OPERATION: Revision, left total knee replacement.

ANESTHESIA: General. TOURNIQUET

TIME: Approximately 80 minutes at 300 mmHg pressure.

DESCRIPTION OF PROCEDURE:

Under satisfactory anesthesia, prophylactic IV antibiotics were given. Sterile prep and drape of the left leg was done in standard manner. The leg was exsanguinated with an Esmarch bandage, and tourniquet around the thigh was inflated to 300 mmHg pressure through the time of bandage application. Anterior approach was performed with a medial parapatellar arthrotomy. There was clear fluid and no clinical sign of infection. The knee had abundant synovitis consistent with third body particle wear. Synovectomy was performed as part of the exposure. The size of the leg and tightness of the extensor mechanism made it difficult to evert the patella, and we performed a quad snip. We mobilized the patella out of the way. The patella and tibia were still solidly fixed to bone. We lifted the rotating platform tray out of the tibia, and noted multiple circumferential wear lines in it. The femoral component was noted to have early loosening with lysis underneath. We were able to easily extract the femoral component. It was posterior stabilized [**Last Name (un) 10**] Sigma PS implant. We then reamed the femoral canal for 120 mm depth up to 14 mm. We did a slight 1-2 mm trim of the distal femur. We placed the 4-in-1 cutting block onto the end of the reamer and set the rotation in a gap balance technique. Minimal additional trims were done for the lateral chamfer cut and anterior cut. Posteriorly, we cut through the 4 mm slots. We placed the final jig and recut the housing. We assembled the trial femoral component with a posterior stabilized 2.5 bearing surface, two, 4 mm posterior augments, a 2-mm offset bolt, a 5-degree adapter, and a 115 x 14 mm fluted stemextension. This was impacted in place and was an excellent fit. We placed a 12.5 thick RP-PS insert. The knee came to full extension, flexed well, was well aligned and was stable throughout range of motion. The trials were removed. Bone surfaces were cleaned with pulse lavage and dried. A few sclerotic areas were drilled for cement interdigitation. The permanent implant components were brought on the operating field and assembled. There were all of same size as noted with the trial. Bone surfaces were dried while cement was mixed. The component was cemented on the underside of the articular component and up the femoral rod just to the flutes. The real component was impacted into place. Excess cement was removed. A 2.5 PS tibial insert with a thickness of 12.5 mm was dropped into the RP tray. Final washout was done. Closure was done with #5 Ethibond to the snip area, #1 Vicryl to the capsule, 2-0 Vicryl to the subcutaneous layer and staples to the skin. Sterile dressing was applied. Tourniquet was deflated. The anesthesia was discontinued. She was taken to the Recovery Room in satisfactory condition. Dr. [**Last Name (STitle) **] was present for the entire surgical procedure. There were no known intraoperative complications.

*3.1.2 Structured data preprocessing and feature extraction:*

We extracted patient information from the record associated with each encounter as explained in Table 2. We used all available diagnosis codes (i.e.: principal diagnosis, secondary diagnosis, and other diagnoses) to extract the cause of patients' admission.

As shown in Figure 4, categorical features are handled with either one-hot encoding or "text representation", allowing the model to select the best alternative that improves the accuracy and performance of predictive models. "Text representation" is the process of converting the categorical variables from structured tables into a textual representation[31]. If the data were time-sensitive such as patient's encounter, we kept the order and append the data accordingly. If the data was not time sensitive such as medications, lab tests, and demographic, only the data is appended without considering time components.

One-hot encoders were also used and assessed for categorical variables in medication, diagnosis, health history, and laboratory tests. For diagnosis features, we mapped any ICD9 codes to ICD10 and retained the first three characters (encoding the subchapter of ICD10 code) to reduce the code sparsity. To encode variables extracted from the health history feature set, we combined one-hot indicators generated from categorical features with numerical values for continuous variables. Similarly, we encoded laboratory tests and medication features using one-hot vectors that indicate whether a patient received or not received a specific test or medication. We dropped the variables with missing values from the patient health history table (as listed in Appendix). Ultimately, for each patient $w$, this extraction and one-hot encoding yielded $j$ encounter records denoted as $(w_1, ..., w_j)$. The encounters are ordered by the clinical encounter date, each containing 4,065 variables ($w_j \in R^{4065}$) associated with the encounter $j$.

### *Clinical Text Processing and Feature Extraction*

Before its use, clinical text is first cleaned and pre-processed, as shown in Figure 4. Each patient unstructured data was associated with a list of $N$ clinical notes $(x_1, ..., x_n)$, denoting the number of distinct patient encounters. For facilitating the analysis, the list is ordered by the clinical encounter date and if the model is designed to utilize structured data, a summary statistic of the structured data is appended to the list. We tokenized each note into a list of words and then lowercase stemmed these words.

We constructed a matrix with the columns representing unique vocabulary of words denoted by $V$. The tokens are then represented by one-hot vectors, building a matrix of $L_n$ rows and $|V|$ columns, where each $x_n \in {}^{Ln \times |V|}$ denotes the sequence of $L_n$ words in note $n$. We simply concatenated all the notes with a special delineating marker, yielding a single note of $L_{concat} \times |V|$, where $L_{concat} = \sum_{i=1}^{N} L_i$.

## A.4 Models Summary results

**Table A.2: Summary results (AUROC) for Independent models using clinical texts only (-- shows non-convergence for a specified historical data timeframe)**

| | **Hip Surgery** | | | | | | | **Knee Surgery** | | | | | | |
|---|---|---|---|---|---|---|---|---|---|---|---|---|---|---|
| | **Only history** | **0** | **3** | **6** | **12** | **24** | **36** | **only history** | **0** | **3** | **6** | **12** | **24** | **36** |
| **Model A** | 0.773 | 0.846 | **0.852** | 0.835 | 0.838 | 0.819 | 0.820 | 0.914 | 0.832 | 0.927 | **0.947** | 0.922 | 0.893 | 0.876 |
| **Model B** | 0.782 | 0.852 | 0.855 | **0.895** | 0.844 | 0.791 | 0.846 | 0.912 | 0.845 | 0.895 | **0.911** | 0.898 | 0.907 | 0.876 |
| **Model C** | 0.807 | 0.855 | 0.876 | **0.878** | 0.845 | 0.814 | 0.846 | 0.911 | 0.858 | **0.917** | 0.908 | 0.893 | 0.906 | 0.876 |
| **Model D** | 0.770 | 0.837 | 0.852 | **0.875** | 0.839 | 0.819 | 0.847 | 0.884 | 0.868 | **0.916** | 0.903 | 0.910 | 0.897 | 0.874 |
| **Model E** | 0.827 | 0.873 | **0.898** | 0.886 | 0.787 | 0.816 | 0.803 | 0.865 | 0.847 | 0.913 | **0.928** | 0.867 | 0.849 | 0.848 |
| **Model F** | 0.763 | 0.789 | **0.865** | 0.757 | 0.686 | -- | -- | 0.856 | 0.648 | **0.892** | 0.866 | 0.832 | -- | -- |
| **Model G** | 0.863 | 0.865 | 0.865 | **0.869** | 0.809 | 0.819 | 0.826 | 0.924 | 0.905 | 0.928 | **0.929** | 0.925 | 0.87 | 0.88 |
| **Model H** | 0.827 | 0.875 | 0.885 | **0.886** | 0.787 | 0.816 | 0.803 | 0.865 | 0.847 | 0.914 | **0.928** | 0.867 | 0.849 | 0.849 |
| **Model I** | 0.763 | 0.789 | **0.865** | 0.757 | 0.686 | -- | -- | 0.856 | 0.648 | **0.892** | 0.866 | 0.832 | -- | -- |
| **Model J** | 0.863 | 0.865 | 0.865 | **0.868** | 0.809 | 0.819 | 0.826 | 0.924 | 0.905 | **0.932** | 0.929 | 0.925 | 0.87 | 0.88 |
| **Model K** | 0.704 | 0.811 | 0.863 | 0.698 | **0.786** | 0.744 | 0.677 | 0.794 | 0.713 | **0.853** | 0.779 | 0.803 | 0.712 | 0.617 |
| **Model L** | 0.724 | 0.827 | 0.866 | 0.701 | **0.786** | 0.746 | 0.673 | 0.816 | 0.701 | **0.851** | 0.777 | 0.804 | 0.713 | 0.627 |
| **Model M** | 0.526 | 0.608 | 0.608 | **0.611** | -- | -- | 0.514 | 0.521 | 0.55 | 0.571 | **0.597** | 0.568 | 0.545 | 0.536 |
| **Model N** | 0.787 | 0.843 | 0.843 | **0.87** | 0.791 | 0.785 | 0.785 | 0.871 | 0.852 | 0.913 | **0.922** | 0.856 | 0.852 | 0.825 |
| **Model O** | 0.732 | **0.847** | 0.798 | 0.755 | 0.686 | 0.694 | 0.702 | 0.873 | 0.81 | **0.878** | 0.866 | 0.832 | 0.795 | 0.805 |
| **Model P** | 0.861 | 0.866 | 0.872 | **0.877** | 0.827 | 0.827 | 0.822 | 0.917 | 0.919 | **0.941** | **0.941** | 0.917 | 0.869 | 0.879 |
| **Model Q** | 0.570 | **0.732** | 0.702 | 0.681 | 0.678 | 0.658 | 0.610 | 0.573 | 0.627 | **0.662** | 0.657 | 0.654 | 0.642 | 0.572 |
| **Model R** | 0.795 | 0.857 | 0.874 | **0.877** | 0.868 | 0.789 | 0.825 | 0.898 | 0.881 | 0.905 | **0.920** | 0.889 | 0.891 | 0.859 |
| **Model S** | 0.741 | 0.855 | **0.860** | 0.649 | 0.800 | 0.827 | 0.826 | 0.900 | 0.867 | 0.896 | **0.901** | 0.897 | 0.893 | 0.550 |
| **Model T** | 0.783 | 0.836 | **0.886** | 0.877 | 0.862 | 0.830 | 0.840 | 0.899 | 0.847 | 0.918 | **0.931** | 0.918 | 0.916 | 0.897 |
| **Model U** | 0.660 | 0.747 | 0.733 | **0.764** | 0.745 | 0.735 | 0.758 | 0.703 | 0.685 | 0.777 | **0.820** | 0.779 | 0.747 | 0.706 |

**Model A:** Average(hs=256), **Model B:** Average(hs=256)+Attention(additive)(hs=128), **Model C:** CNN(hs=64)(kernels=3,5,7,9), **Model D:** CNN(hs=64)(kernels=3,5,7,9)+Attention(additive)(hs=128), **Model E:** LR+BOW+norm=None, **Model F:** LR+BOW+norm=l1, **Model G:** LR+BOW+norm=l2, **Model H:**

LR+BinaryBOW+norm=None, **Model I:** LR+BinaryBOW+norm=l1, **Model J:** LR+BinaryBOW+norm=l2, **Model K:** LR+LDA+norm=None, **Model L:** LR+LDA+norm=l2, **Model M:** LR+Structured, **Model N:** LR+TFIDF+norm=None, **Model O:** LR+TFIDF+norm=l1, **Model P:** LR+TFIDF+norm=l2, **Model Q:** LSTM(hs=128), **Model R:** LSTM(hs=128)+Attention(additive)(hs=128), **Model S:** LSTM(hs=128)+Attention(additive)(hs=128)+LSTM(hs=128)+Attention(additive)(hs=128), **Model T:** maxpoolLSTM(hs=128), **Model U:** BERT

**Table A.3: Summary results (AUROC) for multi-task models using clinical texts only (-- shows non-convergence for a specified historical data timeframe)**

| | **Hip Surgery** | | | | | | | **Knee Surgery** | | | | | | |
|---|---|---|---|---|---|---|---|---|---|---|---|---|---|---|
| **Months** | **Only history** | **0** | **3** | **6** | **12** | **24** | **36** | **Only history** | **0** | **3** | **6** | **12** | **24** | **36** |
| **Model A** | 0.814 | 0.844 | **0.862** | 0.858 | 0.859 | 0.851 | 0.826 | 0.916 | 0.849 | 0.929 | **0.938** | 0.917 | 0.889 | 0.907 |
| **Model B** | 0.804 | 0.867 | **0.890** | 0.862 | 0.845 | 0.851 | 0.816 | 0.894 | 0.864 | 0.921 | **0.922** | **0.922** | 0.895 | 0.899 |
| **Model C** | 0.799 | 0.848 | 0.865 | **0.866** | 0.827 | 0.835 | 0.841 | 0.884 | 0.856 | 0.917 | **0.923** | 0.911 | 0.864 | 0.908 |
| **Model D** | 0.810 | 0.845 | **0.863** | 0.858 | 0.824 | 0.842 | 0.855 | 0.885 | 0.848 | 0.919 | **0.926** | 0.922 | 0.878 | 0.884 |
| **Model E** | 0.802 | 0.791 | 0.784 | **0.836** | 0.787 | 0.816 | 0.803 | 0.844 | 0.797 | 0.883 | **0.908** | 0.867 | 0.869 | 0.878 |
| **Model F** | 0.738 | 0.682 | **0.776** | 0.707 | 0.686 | -- | -- | 0.835 | 0.598 | **0.862** | 0.846 | 0.832 | -- | -- |
| **Model G** | 0.838 | **0.867** | 0.860 | 0.855 | 0.809 | 0.819 | 0.826 | 0.903 | 0.855 | 0.929 | **0.938** | 0.925 | 0.890 | 0.910 |
| **Model H** | 0.802 | 0.791 | 0.784 | **0.836** | 0.787 | 0.816 | 0.803 | 0.844 | 0.797 | 0.884 | **0.908** | 0.867 | 0.869 | 0.879 |
| **Model I** | 0.738 | 0.682 | **0.776** | 0.707 | 0.686 | -- | -- | 0.835 | 0.598 | **0.862** | 0.846 | 0.832 | -- | -- |
| **Model J** | 0.838 | **0.867** | 0.860 | 0.855 | 0.809 | 0.819 | 0.826 | 0.903 | 0.855 | 0.929 | **0.938** | 0.925 | 0.890 | 0.910 |
| **Model K** | 0.679 | 0.704 | 0.774 | 0.648 | **0.786** | 0.744 | 0.677 | 0.773 | 0.663 | **0.823** | 0.759 | 0.803 | 0.732 | 0.647 |
| **Model L** | 0.699 | 0.720 | 0.777 | 0.651 | **0.786** | 0.746 | 0.673 | 0.795 | 0.651 | **0.821** | 0.757 | 0.804 | 0.733 | 0.657 |
| **Model M** | -- | -- | 0.522 | **0.535** | -- | -- | 0.514 | -- | -- | 0.541 | **0.577** | 0.568 | 0.565 | 0.566 |

| **Model N** | 0.762 | 0.736 | 0.781 | **0.834** | 0.791 | 0.785 | 0.785 | 0.850 | 0.802 | 0.883 | **0.902** | 0.856 | 0.872 | 0.855 |
|---|---|---|---|---|---|---|---|---|---|---|---|---|---|---|
| **Model O** | 0.707 | **0.740** | 0.709 | 0.705 | 0.686 | 0.694 | 0.702 | **0.852** | 0.760 | 0.828 | 0.846 | 0.832 | 0.815 | 0.835 |
| **Model P** | 0.836 | 0.849 | 0.852 | **0.859** | 0.827 | 0.827 | 0.822 | 0.896 | 0.869 | 0.927 | **0.933** | 0.917 | 0.889 | 0.909 |
| **Model Q** | 0.548 | **0.709** | 0.682 | 0.700 | 0.656 | 0.656 | 0.625 | **0.625** | 0.622 | 0.602 | 0.603 | 0.566 | 0.623 | 0.604 |
| **Model R** | 0.807 | **0.861** | 0.851 | 0.859 | 0.842 | 0.85 | 0.838 | 0.882 | 0.866 | 0.898 | 0.903 | **0.912** | 0.879 | 0.901 |
| **Model S** | 0.811 | **0.875** | 0.834 | 0.816 | -- | -- | -- | 0.897 | **0.909** | 0.885 | 0.879 | -- | -- | -- |
| **Model T** | 0.790 | 0.848 | **0.878** | 0.860 | 0.850 | 0.853 | 0.855 | 0.920 | 0.871 | 0.928 | **0.933** | 0.922 | 0.896 | 0.912 |

**Model A:** Average(hs=256), **Model B:** Average(hs=256)+Attention(additive)(hs=128), **Model C:** CNN(hs=64)(kernels=3,5,7,9), **Model D:** CNN(hs=64)(kernels=3,5,7,9)+Attention(additive)(hs=128), **Model E:** LR+BOW+norm=None, **Model F:** LR+BOW+norm=l1, **Model G:** LR+BOW+norm=l2, **Model H:** LR+BinaryBOW+norm=None, **Model I:** LR+BinaryBOW+norm=l1, **Model J:** LR+BinaryBOW+norm=l2, **Model K:** LR+LDA+norm=None, **Model L:** LR+LDA+norm=l2, **Model M:** LR+Structured, **Model N:** LR+TFIDF+norm=None, **Model O:** LR+TFIDF+norm=l1, **Model P:** LR+TFIDF+norm=l2, **Model Q:** LSTM(hs=128), **Model R:** LSTM(hs=128)+Attention(additive)(hs=128), **Model S:** LSTM(hs=128)+Attention(additive)(hs=128)+LSTM(hs=128)+Attention(additive)(hs=128), **Model T:** maxpoolLSTM(hs=128)

**Table A.4: Summary results (AUROC) for independent models using structured data and clinical texts (-- shows non-convergence for a specified historical data timeframe)**

| | **Hip Surgery** | | | | | | | **Knee Surgery** | | | | | | |
|---|---|---|---|---|---|---|---|---|---|---|---|---|---|---|
| **Months** | **Only history** | **0** | **3** | **6** | **12** | **24** | **36** | **only history** | **0** | **3** | **6** | **12** | **24** | **36** |
| **Model A** | 0.787 | 0.806 | 0.819 | **0.824** | 0.811 | 0.800 | 0.822 | 0.885 | 0.853 | 0.892 | **0.908** | 0.896 | 0.855 | 0.846 |
| **Model B** | 0.737 | **0.790** | 0.786 | 0.76 | 0.756 | 0.758 | 0.760 | 0.792 | 0.770 | 0.811 | **0.818** | 0.782 | 0.767 | 0.740 |
| **Model C** | 0.816 | **0.861** | 0.847 | 0.84 | 0.841 | 0.809 | 0.829 | 0.884 | 0.862 | 0.898 | **0.914** | 0.882 | 0.878 | 0.855 |
| **Model D** | 0.809 | 0.811 | 0.883 | **0.887** | 0.854 | 0.830 | 0.854 | 0.884 | 0.864 | **0.927** | 0.913 | 0.919 | 0.886 | 0.866 |
| **Model E** | 0.788 | 0.811 | 0.804 | 0.815 | 0.794 | **0.832** | 0.810 | 0.855 | 0.818 | **0.893** | 0.883 | 0.873 | 0.863 | 0.833 |
| **Model F** | 0.665 | **0.668** | 0.643 | 0.660 | 0.630 | 0.623 | 0.614 | 0.681 | 0.682 | 0.658 | **0.689** | 0.678 | 0.668 | 0.662 |
| **Model G** | 0.790 | **0.802** | 0.754 | 0.772 | 0.730 | 0.696 | 0.684 | 0.871 | 0.845 | 0.874 | **0.878** | 0.847 | 0.812 | 0.786 |
| **Model H** | 0.788 | 0.811 | 0.804 | 0.814 | 0.794 | **0.832** | 0.810 | 0.855 | 0.818 | **0.893** | 0.883 | 0.873 | 0.863 | 0.833 |
| **Model I** | **0.668** | **0.668** | 0.643 | 0.660 | 0.630 | 0.624 | 0.614 | 0.681 | 0.682 | 0.658 | **0.689** | 0.678 | 0.668 | 0.662 |
| **Model J** | 0.790 | **0.817** | 0.754 | 0.772 | 0.730 | 0.698 | 0.685 | 0.871 | 0.845 | 0.874 | **0.878** | 0.847 | 0.812 | 0.786 |
| **Model K** | 0.689 | **0.729** | 0.666 | 0.697 | 0.652 | 0.632 | 0.624 | **0.766** | 0.713 | 0.702 | 0.749 | 0.700 | 0.664 | 0.657 |
| **Model L** | 0.700 | **0.734** | 0.675 | 0.701 | 0.660 | 0.636 | 0.624 | **0.760** | 0.715 | 0.707 | 0.749 | 0.708 | 0.664 | 0.654 |
| **Model M** | **0.668** | **0.668** | 0.643 | 0.660 | 0.630 | 0.624 | 0.614 | 0.681 | 0.682 | 0.657 | **0.689** | 0.678 | 0.668 | 0.662 |
| **Model N** | 0.728 | **0.800** | 0.768 | 0.777 | 0.759 | 0.780 | 0.782 | 0.860 | 0.834 | 0.890 | **0.905** | 0.889 | 0.838 | 0.821 |
| **Model O** | **0.668** | **0.668** | 0.643 | 0.660 | 0.630 | 0.624 | 0.615 | 0.685 | 0.682 | 0.658 | **0.689** | 0.678 | 0.668 | 0.662 |
| **Model P** | 0.784 | **0.802** | 0.755 | 0.776 | 0.728 | 0.710 | 0.686 | **0.872** | 0.839 | 0.866 | 0.871 | 0.842 | 0.815 | 0.774 |
| **Model Q** | 0.800 | 0.793 | 0.799 | **0.812** | 0.809 | 0.763 | 0.792 | 0.841 | 0.873 | 0.664 | **0.899** | 0.703 | 0.861 | 0.776 |
| **Model R** | 0.788 | 0.800 | **0.822** | 0.809 | 0.795 | 0.798 | 0.819 | 0.893 | 0.868 | 0.885 | **0.920** | 0.888 | 0.868 | 0.848 |
| **Model S** | 0.681 | **0.774** | 0.736 | 0.733 | 0.718 | 0.717 | 0.701 | 0.684 | 0.702 | **0.722** | 0.718 | 0.714 | 0.687 | 0.672 |
| **Model T** | 0.819 | 0.805 | 0.871 | **0.885** | 0.836 | 0.813 | 0.842 | 0.890 | 0.863 | **0.918** | **0.918** | 0.892 | 0.880 | 0.851 |

**Model A:** Average(hs=256)+Attention(additive)(hs=128)+Structured, **Model B:** Average(hs=256)+Structured, **Model C:** CNN(hs=64)(kernels=3,5,7,9)+Attention(additive)(hs=128)+Structured, **Model D:** CNN(hs=64)(kernels=3,5,7,9)+Structured, **Model E:** LR+BOW+norm=None+Structured, **Model F:** LR+BOW+norm=l1+Structured, **Model G:** LR+BOW+norm=l2+Structured, **Model H:** LR+BinaryBOW+norm=None+Structured, **Model I:** LR+BinaryBOW+norm=l1+Structured, **Model J:** LR+BinaryBOW+norm=l2+Structured, **Model K:** LR+LDA+norm=None+Structured, **Model L:** LR+LDA+norm=l2+Structured, **Model M:** LR+Structured, **Model N:** LR+TFIDF+norm=None+Structured, **Model O:** LR+TFIDF+norm=l1+Structured, **Model P:** LR+TFIDF+norm=l2+Structured, **Model Q:** LSTM(hs=128)+Attention(additive)(hs=128)+LSTM(hs=128)+Attention(additive)(hs=128)+Structured, **Model R:** LSTM(hs=128)+Attention(additive)(hs=128)+Structured, **Model S:** LSTM(hs=128)+Structured, **Model T:** maxpoolLSTM(hs=128)+Structured

**Table A.5: Summary results (AUROC) for multitask models using structured data and clinical texts (-- shows non-convergence for a specified historical data timeframe)**

| | Hip Surgery | | | | | | | Knee Surgery | | | | | | |
|---|---|---|---|---|---|---|---|---|---|---|---|---|---|---|
| **Months** | **Only history** | **0** | **3** | **6** | **12** | **24** | **36** | **Only history** | **0** | **3** | **6** | **12** | **24** | **36** |
| **Model A** | 0.795 | 0.801 | **0.835** | 0.799 | 0.798 | 0.798 | 0.807 | 0.877 | 0.868 | 0.910 | **0.940** | 0.873 | 0.864 | 0.845 |
| **Model B** | 0.776 | 0.795 | 0.818 | 0.793 | **0.828** | 0.795 | 0.788 | 0.812 | 0.773 | **0.841** | 0.824 | 0.826 | 0.744 | 0.764 |
| **Model C** | 0.807 | 0.800 | **0.865** | 0.854 | 0.844 | 0.844 | 0.847 | 0.875 | 0.877 | 0.891 | 0.905 | **0.927** | 0.876 | 0.868 |
| **Model D** | 0.799 | 0.881 | **0.888** | 0.868 | 0.852 | 0.805 | 0.842 | 0.880 | 0.881 | **0.930** | 0.899 | 0.890 | 0.888 | 0.883 |
| **Model E** | 0.785 | 0.790 | 0.791 | **0.817** | 0.781 | 0.800 | 0.794 | 0.839 | 0.813 | 0.876 | **0.896** | 0.848 | 0.861 | 0.852 |
| **Model F** | **0.688** | 0.673 | 0.654 | 0.664 | 0.642 | 0.630 | 0.610 | **0.691** | 0.673 | 0.638 | 0.677 | 0.669 | 0.672 | 0.664 |
| **Model G** | 0.784 | **0.804** | 0.78 | 0.778 | 0.764 | 0.736 | 0.732 | 0.864 | 0.842 | **0.879** | 0.873 | 0.848 | 0.815 | 0.800 |
| **Model H** | 0.785 | 0.790 | 0.791 | **0.817** | 0.780 | 0.800 | 0.794 | 0.839 | 0.813 | 0.876 | **0.897** | 0.848 | 0.861 | 0.852 |
| **Model I** | **0.689** | 0.674 | 0.654 | 0.664 | 0.643 | 0.630 | 0.610 | **0.691** | 0.673 | 0.638 | 0.677 | 0.669 | 0.672 | 0.664 |
| **Model J** | 0.784 | **0.801** | 0.78 | 0.778 | 0.764 | 0.736 | 0.732 | 0.864 | 0.842 | **0.879** | 0.873 | 0.848 | 0.815 | 0.800 |
| **Model K** | 0.683 | **0.738** | 0.683 | 0.690 | 0.721 | 0.638 | 0.618 | 0.721 | 0.693 | 0.711 | 0.698 | **0.722** | 0.666 | 0.662 |
| **Model L** | 0.702 | **0.738** | 0.696 | 0.708 | 0.730 | 0.643 | 0.629 | **0.755** | 0.691 | 0.714 | 0.722 | 0.731 | 0.663 | 0.672 |
| **Model M** | 0.673 | **0.675** | 0.654 | 0.664 | 0.642 | 0.629 | 0.609 | 0.673 | 0.673 | 0.638 | **0.677** | 0.669 | 0.672 | 0.663 |
| **Model N** | 0.747 | 0.765 | 0.774 | **0.845** | 0.756 | 0.808 | 0.761 | 0.846 | 0.816 | **0.899** | 0.876 | 0.858 | 0.857 | 0.817 |
| **Model O** | **0.701** | 0.674 | 0.654 | 0.664 | 0.643 | 0.630 | 0.61 | **0.707** | 0.673 | 0.638 | 0.677 | 0.669 | 0.672 | 0.664 |
| **Model P** | 0.788 | **0.808** | 0.778 | 0.774 | 0.761 | 0.734 | 0.725 | 0.861 | 0.839 | 0.873 | **0.875** | 0.842 | 0.808 | 0.797 |
| **Model Q** | 0.802 | 0.791 | 0.840 | **0.844** | 0.816 | -- | 0.809 | 0.880 | 0.868 | 0.878 | **0.906** | 0.871 | -- | 0.852 |
| **Model R** | 0.805 | 0.789 | **0.853** | 0.852 | **0.853** | 0.812 | 0.834 | 0.878 | 0.874 | 0.880 | **0.908** | 0.888 | 0.831 | 0.836 |
| **Model S** | 0.706 | **0.805** | 0.764 | 0.724 | 0.727 | 0.722 | 0.740 | 0.683 | **0.732** | 0.693 | 0.722 | 0.703 | 0.680 | 0.681 |
| **Model T** | 0.810 | 0.804 | **0.873** | 0.851 | 0.828 | 0.809 | 0.836 | 0.879 | 0.852 | **0.922** | 0.918 | 0.899 | 0.870 | 0.872 |

**Model A:** Average(hs=256)+Attention(additive)(hs=128)+Structured, **Model B:** Average(hs=256)+Structured, **Model C:** CNN(hs=64)(kernels=3,5,7,9)+Attention(additive)(hs=128)+Structured, **Model D:** CNN(hs=64)(kernels=3,5,7,9)+Structured, **Model E:** LR+BOW+norm=None+Structured, **Model F:** LR+BOW+norm=l1+Structured, **Model G:** LR+BOW+norm=l2+Structured, **Model H:** LR+BinaryBOW+norm=None+Structured, **Model I:** LR+BinaryBOW+norm=l1+Structured, **Model J:** LR+BinaryBOW+norm=l2+Structured, **Model K:** LR+LDA+norm=None+Structured, **Model L:** LR+LDA+norm=l2+Structured, **Model M:** LR+Structured, **Model N:** LR+TFIDF+norm=None+Structured, **Model O:** LR+TFIDF+norm=l1+Structured, **Model P:** LR+TFIDF+norm=l2+Structured, **Model Q:** LSTM(hs=128)+Attention(additive)(hs=128)+LSTM(hs=128)+Attention(additive)(hs=128)+Structured, **Model R:** LSTM(hs=128)+Attention(additive)(hs=128)+Structured, **Model S:** LSTM(hs=128)+Structured, **Model T:** maxpoolLSTM(hs=128)+Structured

**Table A.6: Summary results (AUROC) for multitask models using structured data (-- shows non-convergence for a specified historical data timeframe)**

| | **Hip Surgery** | | | | | | | **Knee Surgery** | | | | | | |
|---|---|---|---|---|---|---|---|---|---|---|---|---|---|---|
| **Months** | **Only history** | **0** | **3** | **6** | **12** | **24** | **36** | **Only history** | **0** | **3** | **6** | **12** | **24** | **36** |
| **Model A** | 0.601 | 0.603 | 0.672 | 0.666 | 0.684 | **0.718** | 0.707 | 0.623 | 0.623 | 0.654 | **0.681** | 0.672 | 0.662 | 0.643 |
| **Model B** | 0.568 | 0.568 | 0.66 | 0.660 | 0.678 | 0.676 | **0.700** | 0.669 | 0.670 | 0.673 | 0.680 | **0.753** | 0.729 | 0.715 |
| **Model C** | 0.606 | 0.643 | 0.692 | 0.666 | 0.700 | 0.702 | **0.723** | 0.648 | 0.648 | 0.660 | 0.684 | **0.720** | 0.700 | 0.702 |
| **Model D** | 0.601 | 0.604 | 0.672 | 0.658 | 0.684 | 0.722 | **0.730** | 0.623 | 0.623 | 0.654 | **0.684** | 0.672 | 0.661 | 0.641 |
| **Model E** | 0.568 | 0.568 | 0.656 | 0.660 | 0.678 | 0.676 | **0.700** | 0.669 | 0.669 | 0.693 | 0.654 | 0.737 | **0.729** | 0.715 |
| **Model F** | 0.606 | 0.606 | 0.692 | 0.667 | 0.700 | 0.702 | **0.723** | 0.648 | 0.648 | 0.660 | 0.684 | **0.720** | 0.700 | 0.702 |
| **Model G** | 0.505 | 0.531 | 0.595 | 0.614 | 0.623 | 0.611 | **0.695** | 0.606 | 0.606 | 0.646 | 0.665 | 0.731 | 0.710 | **0.744** |
| **Model H** | 0.503 | 0.529 | 0.597 | 0.599 | 0.611 | 0.618 | **0.663** | 0.598 | 0.592 | 0.664 | 0.656 | 0.735 | 0.710 | **0.739** |
| **Model I** | -- | -- | 0.522 | **0.535** | -- | -- | -- | -- | -- | 0.541 | **0.577** | 0.568 | 0.565 | 0.566 |
| **Model J** | 0.593 | 0.588 | 0.652 | 0.654 | **0.714** | 0.697 | 0.676 | 0.623 | 0.624 | 0.620 | 0.624 | 0.618 | **0.629** | 0.610 |
| **Model K** | 0.558 | 0.558 | 0.657 | 0.661 | 0.698 | 0.694 | **0.727** | 0.661 | 0.661 | 0.665 | 0.684 | **0.706** | 0.696 | 0.693 |
| **Model L** | 0.572 | 0.572 | 0.641 | 0.685 | 0.714 | 0.720 | **0.725** | 0.639 | 0.639 | 0.662 | 0.679 | **0.712** | 0.689 | 0.690 |

**Model A:** LR+BOW+norm=None, **Model B:** LR+BOW+norm=l1, **Model C:** LR+BOW+norm=l2, **Model D:** LR+BinaryBOW+norm=None, **Model E:** LR+BinaryBOW+norm=l1, **Model F:** LR+BinaryBOW+norm=l2, **Model G:** LR+LDA+norm=None, **Model H:** LR+LDA+norm=l2, **Model I:** LR+Structured, **Model J:** LR+TFIDF+norm=None, **Model K:** LR+TFIDF+norm=l1, **Model L:** LR+TFIDF+norm=l2

**Table A.7: Summary results (AUROC) for independent models using structured data (-- shows non-convergence for a specified historical data timeframe)**

| | **Hip Surgery** | | | | | | | **Knee Surgery** | | | | | | |
|---|---|---|---|---|---|---|---|---|---|---|---|---|---|---|
| **Months** | **Only history** | **0** | **3** | **6** | **12** | **24** | **36** | **Only history** | **0** | **3** | **6** | **12** | **24** | **36** |
| **Model A** | 0.611 | 0.607 | 0.609 | 0.666 | 0.689 | **0.706** | **0.706** | 0.623 | 0.623 | 0.654 | **0.690** | 0.718 | 0.662 | 0.643 |
| **Model B** | 0.571 | 0.568 | 0.600 | 0.660 | 0.678 | 0.676 | **0.700** | 0.669 | 0.650 | 0.657 | 0.680 | 0.701 | 0.717 | 0.715 |
| **Model C** | 0.610 | 0.611 | 0.611 | 0.666 | 0.669 | 0.702 | **0.706** | 0.648 | 0.648 | 0.652 | 0.684 | 0.720 | 0.700 | 0.702 |
| **Model D** | 0.608 | 0.604 | 0.608 | 0.658 | 0.689 | **0.706** | 0.701 | 0.623 | 0.623 | 0.654 | **0.690** | 0.728 | 0.661 | 0.641 |
| **Model E** | 0.569 | 0.568 | 0.576 | 0.660 | 0.678 | 0.676 | **0.700** | 0.669 | 0.653 | 0.651 | 0.654 | 0.728 | **0.718** | 0.715 |
| **Model F** | 0.607 | 0.606 | 0.609 | 0.666 | 0.669 | **0.702** | **0.702** | 0.648 | 0.653 | 0.601 | 0.684 | **0.720** | 0.700 | 0.702 |
| **Model G** | 0.508 | 0.539 | 0.595 | 0.614 | 0.623 | 0.611 | **0.695** | 0.606 | 0.606 | 0.646 | 0.665 | 0.722 | 0.710 | **0.721** |
| **Model H** | 0.506 | 0.529 | 0.597 | 0.599 | 0.611 | 0.618 | **0.663** | 0.598 | 0.592 | 0.653 | 0.656 | 0.722 | 0.710 | **0.721** |
| **Model I** | -- | -- | 0.522 | **0.535** | -- | -- | -- | -- | -- | 0.541 | **0.577** | 0.568 | 0.565 | 0.566 |
| **Model J** | 0.595 | 0.589 | 0.611 | 0.654 | 0.669 | **0.697** | 0.676 | 0.623 | 0.624 | 0.620 | 0.624 | 0.618 | **0.629** | 0.610 |
| **Model K** | 0.558 | 0.559 | 0.607 | 0.661 | 0.689 | 0.694 | **0.700** | 0.661 | 0.653 | 0.652 | 0.684 | **0.706** | 0.696 | 0.693 |
| **Model L** | 0.572 | 0.585 | 0.598 | 0.645 | 0.674 | **0.702** | 0.701 | 0.639 | 0.639 | 0.647 | 0.679 | **0.712** | 0.689 | 0.690 |

**Model A:** LR+BOW+norm=None, **Model B:** LR+BOW+norm=l1, **Model C:** LR+BOW+norm=l2, **Model D:** LR+BinaryBOW+norm=None, **Model E:** LR+BinaryBOW+norm=l1, **Model F:** LR+BinaryBOW+norm=l2, **Model G:** LR+LDA+norm=None, **Model H:** LR+LDA+norm=l2, **Model I:** LR+Structured, **Model J:** LR+TFIDF+norm=None, **Model K:** LR+TFIDF+norm=l1, **Model L:** LR+TFIDF+norm=l2

## References

1 Hudson, K., Lifton, R. & Patrick-Lake, B. The precision medicine initiative cohort program—Building a Research Foundation for 21st Century Medicine. *Precision Medicine Initiative (PMI) Working Group Report to the Advisory Committee to the Director, ed* (2015).

2 JaWanna Henry, M., Yuriy Pylypchuk, P., Talisha Searcy, M., MA; & Vaishali Patel, P. M. *Adoption of Electronic Health Record Systems among U.S. Non-Federal Acute Care Hospitals: 2008-2015*, <https://dashboard.healthit.gov/evaluations/data-briefs/non-federal-acute-care-hospital-ehr-adoption-2008-2015.php#citation> (2016).

3 Dinov, I. D. Volume and value of big healthcare data. *Journal of medical statistics and informatics* **4** (2016).

4 Diallo, G. *Some contributions on large scale heterogeneous data and knowledge integration: application to the healthcare domain*, university of bordeaux, (2019).

5 Zolbanin, H. M. & Delen, D. Processing electronic medical records to improve predictive analytics outcomes for hospital readmissions. *Decision Support Systems* **112**, 98-110, doi:https://doi.org/10.1016/j.dss.2018.06.010 (2018).

6 Chen, M., Hao, Y., Hwang, K., Wang, L. & Wang, L. Disease prediction by machine learning over big data from healthcare communities. *Ieee Access* **5**, 8869-8879 (2017).

7 Weber, G. M., Mandl, K. D. & Kohane, I. S. Finding the Missing Link for Big Biomedical Data. *JAMA* **311**, 2479-2480, doi:10.1001/jama.2014.4228 (2014).

8 Kong, H. J. Managing Unstructured Big Data in Healthcare System. *Healthcare informatics research* **25**, 1-2, doi:10.4258/hir.2019.25.1.1 (2019).

9 Juhn, Y. & Liu, H. Artificial intelligence approaches using natural language processing to advance EHR-based clinical research. *Journal of Allergy and Clinical Immunology* **145**, 463-469 (2020).

10 Desai, N. R. *et al.* Association between hospital penalty status under the hospital readmission reduction program and readmission rates for target and nontarget conditions. *Jama* **316**, 2647-2656 (2016).

11 Service, S. H. N. Unplanned hospital readmissions remain a problem in Mass. *WBJ* (2016).

12 Lette, J. *et al.* Artificial intelligence versus logistic regression statistical modelling to predict cardiac complications after noncardiac surgery. *Clinical Cardiology* **17**, 609-614, doi:10.1002/clc.4960171109 (1994).

13 Ben-Assuli, O. & Padman, R. Analysing repeated hospital readmissions using data mining techniques. *Health Systems* **7**, 166-180, doi:10.1080/20476965.2018.1510040 (2018).

14 Clair, A. J. *et al.* Cost analysis of total joint arthroplasty readmissions in a bundled payment care improvement initiative. *The Journal of arthroplasty* **31**, 1862-1865 (2016).

15 Zawadzki, N. *et al.* Readmission due to infection following total hip and total knee procedures: A retrospective study. *Medicine* **96** (2017).

16 Arefjev, S., Fernandez-Rocha, L., Skycak, J. & Stormont, D. Improving Readmission Prediction by Extracting Relevant Information from Clinical Notes.

17 Bacchi, S. *et al.* Prediction of general medical admission length of stay with natural language processing and deep learning: a pilot study. *Internal and Emergency Medicine*, doi:10.1007/s11739-019-02265-3 (2020).

18 Chen, J. H., Alagappan, M., Goldstein, M. K., Asch, S. M. & Altman, R. B. Decaying relevance of clinical data towards future decisions in data-driven inpatient clinical order sets. *International journal of medical informatics* **102**, 71-79 (2017).

19 Kansagara, D. *et al.* Risk Prediction Models for Hospital Readmission: A Systematic Review. *JAMA* **306**, 1688-1698, doi:10.1001/jama.2011.1515 (2011).

20 Zhou, H., Della, P. R., Roberts, P., Goh, L. & Dhaliwal, S. S. Utility of models to predict 28-day or 30-day unplanned hospital readmissions: an updated systematic review. *BMJ open* **6**, e011060, doi:10.1136/bmjopen-2016-011060 (2016).

21 Soguero-Ruiz, C. *et al.* Data-driven Temporal Prediction of Surgical Site Infection. *AMIA ... Annual Symposium proceedings. AMIA Symposium* **2015**, 1164-1173 (2015).

22 FitzHenry, F. *et al.* Exploring the frontier of electronic health record surveillance: the case of postoperative complications. *Medical care* **51**, 509-516, doi:10.1097/MLR.0b013e31828d1210 (2013).

23 Huang, K., Altosaar, J. & Ranganath, R. Clinicalbert: Modeling clinical notes and predicting hospital readmission. *arXiv preprint arXiv:1904.05342* (2019).

24 Craig, E., Arias, C. & Gillman, D. Predicting readmission risk from doctors' notes. *arXiv preprint arXiv:1711.10663* (2017).

25 Liu, X. *et al.* in *2019 IEEE International Conference on Bioinformatics and Biomedicine (BIBM).* 2642-2648 (IEEE).

26 Chen, J. H. & Asch, S. M. Machine learning and prediction in medicine—beyond the peak of inflated expectations. *The New England journal of medicine* **376**, 2507 (2017).

27 Wallace, B. C., Small, K., Brodley, C. E. & Trikalinos, T. A. in *2011 IEEE 11th international conference on data mining.* 754-763 (IEEE).

28 Martin, G. M. Patient education: Total knee replacement (Beyond the Basics).

29 Edelstein, A. I. *et al.* Can the American College of Surgeons risk calculator predict 30-day complications after knee and hip arthroplasty? *The Journal of arthroplasty* **30**, 5-10 (2015).

30 Mesko, N. W. *et al.* Thirty-day readmission following total hip and knee arthroplasty–a preliminary single institution predictive model. *The Journal of arthroplasty* **29**, 1532-1538 (2014).

31 Nguyen, P., Tran, T., Wickramasinghe, N. & Venkatesh, S. $\mathtt {Deepr} $: a convolutional net for medical records. *IEEE journal of biomedical and health informatics* **21**, 22-30 (2016).

32 Zhang, Y., Jin, R. & Zhou, Z.-H. Understanding bag-of-words model: a statistical framework. *International Journal of Machine Learning and Cybernetics* **1**, 43-52 (2010).

33 JOACHIMS, T. A Probabilistic Analysis of the Rocchio Algorithm with TFIDF for Text Categorization(Topical Report). (1996).

34 Blei, D. M., Ng, A. Y. & Jordan, M. I. Latent dirichlet allocation. *Journal of machine Learning research* **3**, 993-1022 (2003).

35 Bishop, C. M. *Neural networks for pattern recognition*. (Oxford university press, 1995).

36 Hubel, D. H. & Wiesel, T. N. Receptive fields and functional architecture of monkey striate cortex. *The Journal of physiology* **195**, 215-243 (1968).

37 Hochreiter, S. & Schmidhuber, J. Long short-term memory. *Neural computation* **9**, 1735-1780 (1997).

38 Vaswani, A. *et al.* in *Advances in neural information processing systems.* 5998-6008.

39 Yang, Z. *et al.* in *Proceedings of the 2016 conference of the North American chapter of the association for computational linguistics: human language technologies.* 1480-1489.

40 Devlin, J., Chang, M.-W., Lee, K. & Toutanova, K. Bert: Pre-training of deep bidirectional transformers for language understanding. *arXiv preprint arXiv:1810.04805* (2018).

41 Alsentzer, E. *et al.* Publicly available clinical BERT embeddings. *arXiv preprint arXiv:1904.03323* (2019).

42 Caruana, R. Multitask learning. *Machine learning* **28**, 41-75 (1997).